%% file: main.tex
\documentclass[conference]{IEEEtran}
\input{settings/macros}

\IEEEoverridecommandlockouts

\def\BibTeX{{\rm B\kern-.05em{\sc i\kern-.025em b}\kern-.08em
    T\kern-.1667em\lower.7ex\hbox{E}\kern-.125emX}}

\title{Inverse Cross-spectral Neural Networks for Multivariate Time Series\\
\thanks{The work was supported by the SNS JU project 6G-GOALS under the EU’s Horizon program Grant Agreement No 101139232, by Huawei Technology France SASU under Grant N. Tg20250616041, and by the  European Union under the Italian National Recovery and Resilience Plan of NextGenerationEU, partnership on Telecommunications of the Future (PE00000001- program RESTART).}
}

\author{
\IEEEauthorblockN{
Lorenzo Marinucci\IEEEauthorrefmark{1},
Leonardo Di Nino\IEEEauthorrefmark{2},
Gabriele D'Acunto\IEEEauthorrefmark{2},\\
Paolo Di Lorenzo\IEEEauthorrefmark{2},
Sergio Barbarossa\IEEEauthorrefmark{2}
}

\IEEEauthorblockA{
\IEEEauthorrefmark{1}Statistical Sciences Dept., Sapienza University of Rome, Italy \;\\
\IEEEauthorrefmark{2}DIET  Dept., Sapienza University of Rome, Italy \;\\
}

\IEEEauthorblockA{
\{l.marinucci,
leonardo.dinino,
gabriele.dacunto,
paolo.dilorenzo,
sergio.barbarossa\}@uniroma1.it
}
}
\begin{document}
\maketitle

\begin{abstract}
CoVariance Neural Networks and their extensions have emerged as effective tools for processing multivariate data, deriving graph shift operators directly from second-order statistics. These architectures, however, are designed for independent and identically distributed observations and do not fully capture the joint structure of temporal and cross-variable dependencies in multivariate time series. In this work, we introduce Inverse Cross-Spectral Neural Networks (iCSNNs), a class of graph neural networks for stationary multivariate time series whose shift operators are the inverse cross-spectral density (iCSD) matrices. These operators encode frequency-specific conditional relationships among variables, exploiting the decomposition provided by the spectral representation theorem. Leveraging spectral smoothness, frequencies are grouped into bands sharing a single iCSD operator, yielding a compact parametrisation that retains the frequency-dependent structure of the process. We further propose a joint learning procedure to estimate both the Fourier-domain dependence structure and the iCSNN parameters, adapting the iCSD operators to the downstream task. When tested on synthetic data, iCSNN outperforms baselines from different methodological families.

\end{abstract}

\input{sections/Introduction}

\input{sections/background}
\input{sections/ICSNN}

\input{sections/problem_formulation}
\input{sections/problem_solution}

\input{sections/numerical_results}

\input{sections/conclusions}
\bibliographystyle{unsrt}
\bibliography{bibliography}

\end{document}

%% file: settings/macros.tex
\usepackage[english]{babel}
\usepackage[dvipsnames]{xcolor}
\usepackage{stfloats}
\usepackage{type1cm} % type1 computer modern font
\usepackage{graphicx} % advanced figures
\usepackage{xspace} % fix space in macros
\usepackage{balance} % to better equalize the last page
\usepackage{booktabs} % nicer tables
\usepackage{multirow} % multi rows for tables
\usepackage[font={bf}, tableposition=top]{caption} % captions on top for tables
\usepackage{subcaption} % subfloats
\usepackage{bold-extra} % bold + {small capital, italic}
\usepackage{algorithm}
\usepackage{algorithmic}
\usepackage{microtype} % compress text
\usepackage{siunitx} % \num for decimal grouping
\usepackage{xfrac} % nicer slanted fractions
\usepackage{amsmath}
\usepackage{amssymb}
\usepackage{mathtools}
\usepackage{amsfonts}
\usepackage{amsthm}
\usepackage{bbm}
\usepackage[hyphens]{url} % handle long urls
\usepackage[bookmarks, pdftex, colorlinks=true]{hyperref} % clickable references
\usepackage{cleveref} % smart references
\usepackage{csquotes}
\usepackage{bm}
\usepackage{xspace}
\usepackage{physics}
\usepackage{newunicodechar}
\newunicodechar{⟓}{\ensuremath{\uplus}}
\usepackage{tikz}
\usetikzlibrary{positioning}
\usetikzlibrary{bayesnet}
\usetikzlibrary{arrows}
\usetikzlibrary{shapes}
\usetikzlibrary{fit}
\usepackage{tikz-cd}
\usepackage{bbold}
\usetikzlibrary{calc, positioning, arrows.meta, shapes.geometric, fit, backgrounds}
\usepackage{pgf}
\usepackage{tikz-3dplot}
\usepackage{tcolorbox}
\usepackage{pgfplots}
\pgfplotsset{compat=1.18}

\newcommand{\spara}[1]{\smallskip\noindent\textbf{#1}}

\hypersetup{
   linkcolor={red!50!black},
   filecolor={green!50!black},
   citecolor={green!50!black}, 
   urlcolor={blue!80!black},
}

\graphicspath{{fig}}

\theoremstyle{plain}

\theoremstyle{remark}
\newtheorem{remark}{Remark}

\crefname{theorem}{Thm.}{Thms.}
\crefname{proposition}{Prop.}{Props.}
\crefname{lemma}{lem.}{lems.}
\crefname{corollary}{Cor.}{Cors.}
\crefname{definition}{Def.}{Defs.}
\crefname{section}{Sec.}{Secs.}
\crefname{figure}{Fig.}{Figs.}
\crefname{problem}{Prob.}{Probs.}
\crefname{appendix}{App.}{Apps.}
\crefname{equation}{Eq.}{Eqs.}
\crefname{table}{Tab.}{Tabs.}

\definecolor{mypurple}{RGB}{254, 68, 218}

%% file: sections/Introduction.tex
\section{Introduction}
Learning from multivariate time series data is central in many applications, including
neuroscience, finance and climate science, %and sensor networks, 
where the goal is
often to predict labels, continuous responses, or future trajectories from
multiple interacting signals \cite{bullmore2009complex, tsay2013multivariate, Runge2019}. A large amount of work has therefore developed
effective predictive models for classification, regression, and forecasting,
ranging from distance-based and feature-based methods to convolutional-kernel
and graph neural networks (GNNs) %deep neural architectures 
\cite{ding2008querying, tan2022multirocket, GNN_on_timeseries}. In particular, GNNs extract representations by propagating and combining information across variables according to an underlying graph.

In many applications, however, the graph topology is not known a priori and must be learned from data. A first approach \cite{VNNs} addresses this problem by extracting the graph directly from the statistical description of the data. 
The GNN shift operator is set to the sample covariance matrix, and the method is known as CoVariance Neural Network (VNN). 
This construction inherits the expressive power and transferability of polynomial graph filters and
recovers the representational space of Principal Component Analysis (PCA) \cite{transferability_VNN}. Pushing in the same direction,
Precision Neural Networks (PNNs) \cite{cavallo2026pnn} replace the covariance with the precision matrix.
In this way, PNNs operate on conditional rather than marginal dependence, exploiting the sparsity of the precision.

Despite these developments, existing covariance-based methods do not explicitly model cross-temporal dependencies. An extension for time series forecasting is provided by Spatiotemporal Covariance Neural Networks (STVNNs), which combine covariance-based graph filters with temporal convolutions and update both the covariance estimate and the network parameters online \cite{Cavallo_2024}. Although temporal convolutions account for past observations, the graph operator remains an estimate of the lag-zero covariance matrix and therefore describes only marginal, contemporaneous relationships among variables. For stationary time series, a complete description of the second-order structure
at all temporal lags is provided by the cross-spectral density (CSD) matrices,
the Fourier transform of the autocovariance sequence \cite{brockwell1991time}. Additionally, their inverses, the inverse cross-spectral density (iCSD) matrices, encode
frequency-specific conditional dependence, in analogy with the precision matrices in the i.i.d. setting \cite{dahlhaus2000graphical}. A broad literature addresses their
estimation from finite records \cite{Bach2004,tugnait2022sparse}, and the resulting
family of operators admits a graphical representation in terms of
multi-frequency partial correlation graphs \cite{DAcunto2024}.

\spara{Contribution.} In this work we introduce Inverse Cross-Spectral Neural Networks
(iCSNNs), a class of GNNs for stationary multivariate time series whose shift
operators are the iCSD matrices. This
architecture builds on the spectral representation, where the statistical
dependence factorizes into a collection of frequency-specific interactions among
the component series. Exploiting the smoothness of the spectral density, the
frequencies are grouped into bands over which a single iCSD operator is
retained, so that the process is described by a small number of band-wise
operators. This substantially reduces parameters while preserving the dependence
structure of the original process. We further propose a joint learning procedure to estimate both the dependence
structure in the Fourier domain and the iCSNN parameters. The statistical
component of the objective builds on established sparse spectral estimation
methods, which recover the iCSD matrices from finite records through a penalised
Whittle likelihood \cite{tugnait2022sparse}, while the supervised term makes the
estimated operators aware of the downstream task. This is particularly relevant in limited-data regimes, where iCSD estimates may exhibit high variance and the supervised term guides the learning process toward the representations most relevant for prediction. Synthetic experiments on a classification task show that iCSNNs outperform baselines drawn from different families, namely
covariance-based architectures \cite{cavallo2026pnn}, convolutional-kernel classifiers \cite{tan2022multirocket}, distance-based
methods and marginal spectral features \cite{ding2008querying}, as well as a disjoint ablation in which
the operators are estimated from the likelihood alone in low data regime.

%% file: sections/background.tex
\section{Background}

Consider a zero-mean multivariate stochastic process
$\{X(t) = [x_1(t),\dots,x_N(t)]^\top \in \mathbb{R}^N : t \in \mathbb{Z}\}$.
We assume that $X(t)$ is weakly stationary, so that its
autocovariance function
\[
    C_\tau = \mathbb{E}\!\left[X(t+\tau)X(t)^\top\right]
\]
depends only on the lag $\tau \in \mathbb{Z}$. 
We further assume that
$\sum_{\tau \in \mathbb{Z}} \|C_\tau\| < \infty$. 

\spara{Cross-spectral density.} Under these assumptions, for each frequency $f \in [0,1)$, the
\textit{cross-spectral density} (CSD) matrix is defined as the Fourier transform of the
autocovariance sequence,
\[
    \Gamma_f =
    \sum_{\tau=-\infty}^{\infty} C_\tau e^{-i2\pi f \tau}
    \in \mathbb{C}^{N \times N}.
\]
We assume that $\Gamma _f$ is nonsingular for every $f \in [0,1)$ and define the inverse cross-spectral density (iCSD) matrix as $\Theta _f = \Gamma _f^{-1}$. By the spectral representation theorem, $X(t)$ admits a frequency-domain representation in terms of orthogonal spectral increments \cite{brockwell1991time}. Under this representation, second-order dependencies are factorized over frequencies, with the CSD matrix $\Gamma_f$ describing marginal dependencies at frequency $f$, and the iCSD matrix $\Theta_f$ encoding the corresponding conditional dependence structure. This allows the linear dependence structure of the process to be analyzed frequency by frequency. Since $C_{-\tau} = C_\tau^\top$, both $\Gamma_f$ and $\Theta_f$ are Hermitian
for every $f$. Additionally, because the process is real-valued,
$\Gamma_{1-f} = \Gamma_f^{*}$ and $\Theta_{1-f} = \Theta_f^{*}$,
so that it suffices to consider $f \in [0,\tfrac{1}{2}]$.

\spara{Partial correlation graphs.}
The family $\{\Theta_f\}_{f\in[0,\tfrac{1}{2}]}$ plays, for stationary multivariate
processes, a role analogous to the precision matrix in the i.i.d.
multivariate setting. In particular, for $i\neq j$, the component processes
$\{x_i(t)\}$ and $\{x_j(t)\}$ are conditionally uncorrelated given the
remaining ones if and only if
$[\Theta_f]_{ij}=0,$ for all  $f\in[0,\tfrac{1}{2}]$ \cite{dahlhaus2000graphical}.
However, we retain a frequency-band resolution of these relations.
Following \cite{DAcunto2024}, let
$\{I_\nu\}_{\nu=1}^{M}$ be a partition of $[0,\tfrac12]$
into contiguous intervals. Note that, under the stated assumptions, the mapping $f \mapsto \Theta_f$ is continuous \cite{brockwell1991time}. Hence, for a sufficiently narrow partition, we assume $\Theta_f$ to be approximately constant over each band $I_\nu$. 
Consequently, for $\nu=1,...,M$, we define a \textit{band-specific partial correlation graph} (PCG) $\mathcal{G}_\nu=(\mathcal{V},\mathcal{E}_\nu)$,
as
\[
    (i,j)\notin\mathcal{E}_\nu
    \quad\Longleftrightarrow\quad
    [\Theta_{f_\nu}]_{ij}=0 
\]
where $f_\nu\in I_\nu$ is a representative frequency.

%% file: sections/ICSNN.tex
\begin{figure}[t]
    \centering
    \includegraphics[width=\columnwidth]{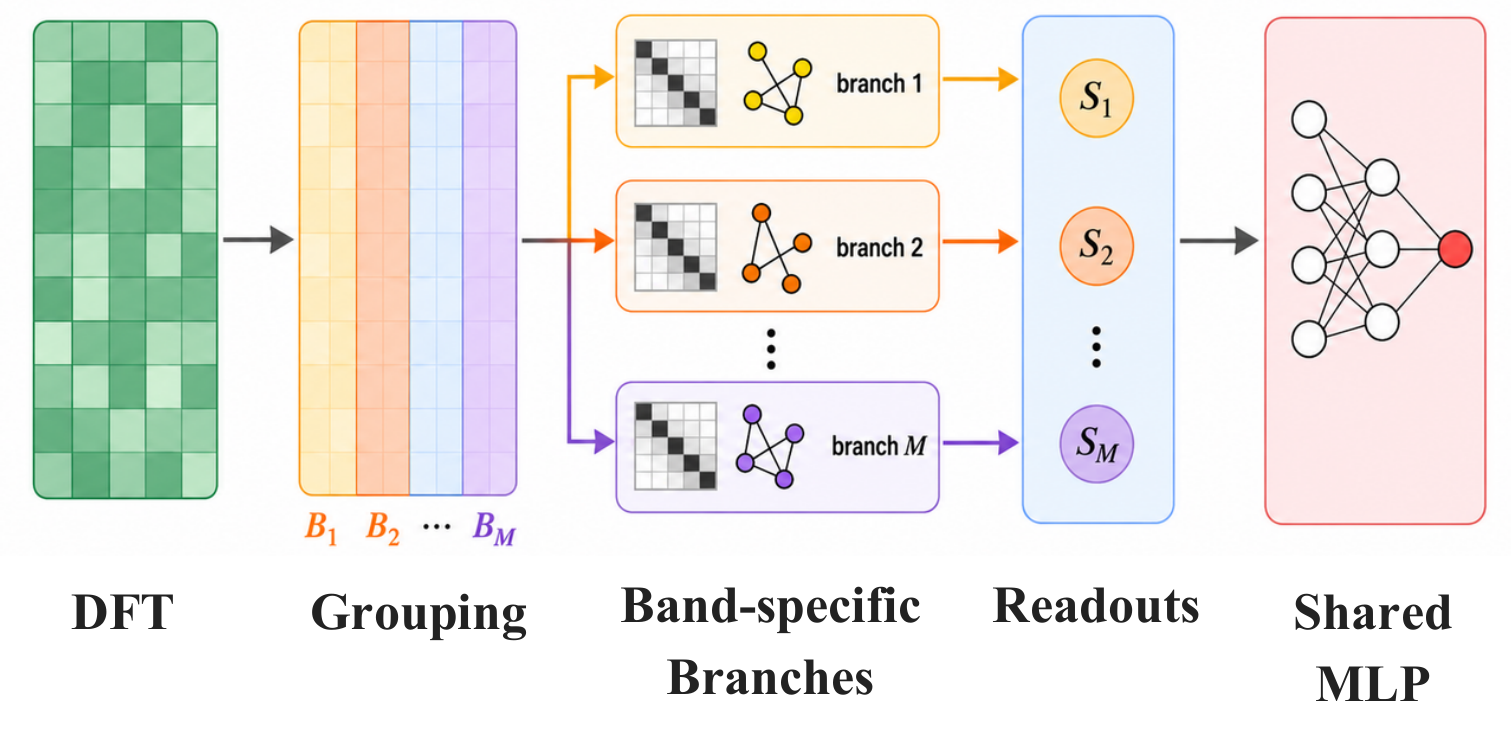}
    \caption{The pipeline of the iCSNN.}
    \label{fig:icsnn}
\end{figure}

\section{Inverse Cross-Spectral Neural Networks}
We consider a supervised learning task in which a multivariate time series $\mathbf{X}\in\mathbb{R}^{N\times T}$, containing $T$ observations of a zero-mean weakly stationary process, is associated with a target $y\in\mathcal{Y}$. We introduce the \textit{Inverse Cross-Spectral Neural Network (iCSNN)}, which maps $\mathbf{X}$ to a prediction $\hat{y}$ by processing its spectral representation in parallel across frequency bands. Applying the one-sided DFT along the temporal dimension yields $\widetilde{\mathbf{X}}\in\mathbb{C}^{N\times F}$, where $F=\lfloor T/2\rfloor+1$.

\spara{Frequency grouping.} The $F$ frequencies are partitioned into $M\ll F$
contiguous bands $\{\mathcal{B}_\nu\}_{\nu=1}^M$ of equal size $B$, and we denote by
$\tilde{\mathbf{X}}_\nu\in\mathbb{C}^{N\times B}$ the Fourier
coefficients falling in band $\nu$. Consistently with the smoothness of the
spectral density, we assume the iCSD to be approximately constant within each
band, so that a single %Hermitian positive definite 
iCSD matrix
$\Theta_{f_\nu}\in\mathbb{C}^{N\times N}$, with $f_\nu \in \mathcal{B}_\nu$, models the conditional second-order
dependence over the band. With a slight abuse of notation, we denote the representative iCSD matrix $\Theta_{f_\nu}$ by $\Theta_\nu$. This reduces the number of iCSD matrices from $F$ to
$M$.

\spara{iCSD filtering.} Each band is processed by an independent branch that
uses $\Theta_{\nu}$ as a GNN 
shift operator of the band-specific PCG $\mathcal{G}_\nu$. A branch stacks $L$
layers, each applying a polynomial filter followed by a complex nonlinearity:
\[
    \tilde{\mathbf{X}}_\nu^{(\ell+1)}
    =
    \sigma\!\left(
        \sum_{k=0}^{K} h_{k,\nu}^{(\ell)}\,\Theta_{\nu}^{k}\,
        \tilde{\mathbf{X}}_\nu^{(\ell)}
    \right),
    \qquad
    \tilde{\mathbf{X}}_\nu^{(0)}=\tilde{\mathbf{X}}_\nu,
\]
with complex coefficients $h_{k,\nu}^{(\ell)}\in\mathbb{C}$, filter order $K$,
and $\sigma$ the modReLU nonlinearity. Since the filter is a polynomial in
$\Theta_{\nu}$, it acts on the eigenvalues of $\Theta_{\nu}$, learning a complex
response along its eigen-directions.

\begin{remark}
    This parallels covariance-based filtering, where covariance operators allow to recover
PCA-type processing along principal directions \cite{VNNs}. Moreover, the sparsity of $\Theta_{\nu}$ yields efficient filtering, by restricting the propagation to the edges of the band-specific PCG $\mathcal{G}_\nu$, in analogy with the PNN approach~\cite{cavallo2026pnn}.
\end{remark}

\spara{Readout and prediction.} After $L$ layers, each branch is reduced to a
real, band-level feature vector. A learnable readout
$W_\nu\in\mathbb{C}^{N\times Q}$, with $Q \in \mathbb{N}$, projects over the node dimension, giving
\[Y_\nu=W_\nu^{H}\tilde{\mathbf{X}}_\nu^{(L)}\in\mathbb{C}^{Q\times B},\]
and a Bartlett-weighted power average over the band produces
\[
    [S_\nu]_i
    =
    \sum_{f\in\mathcal{B}_\nu} a_f\,\big|[Y_\nu]_{if}\big|^2
    \in\mathbb{R}_+,
    \qquad i=1,\ldots,Q.
\]
where the fixed weights $a_f\ge 0$, shared across bands, follow a triangular
(Bartlett) profile,
$a_f \propto 1 - \big|\tfrac{2f-(B-1)}{B-1}\big|$, $f=0,\dots,B-1$, 
peaked at the band centre \cite{brockwell1991time}. This produces real-valued, band-specific features that
summarize the filtered spectral power within each band. 
Finally, a shared MLP maps the concatenated band features $\{S_\nu\}_{\nu=1}^{M}$ to the output:\[\hat{y}=\mathrm{MLP}\big([S_1,\dots,S_M];\,\Omega_{\mathrm{MLP}}\big), \] 
where $\Omega_{\mathrm{MLP}}$ denotes the trainable parameters of the MLP.
Overall, the iCSNN defines a map
$\hat{y}=\Psi(\tilde{\mathbf{X}};\boldsymbol{\Theta},\mathbf{H})$, that depends on
the bandwise iCSD
matrices $\boldsymbol{\Theta}=\{\Theta_{\nu}\}_{\nu=1}^M$, and
the trainable parameters, i.e.\ the filter coefficients, the readouts,
and the MLP weights, $\mathbf{H}=\{h_{k,\nu}^{(\ell)}\}\cup\{W_\nu\}\cup\Omega_{\mathrm{MLP}}.$

%% file: sections/problem_formulation.tex
\section{Problem Formulation}
The iCSNN architecture requires the band-level iCSD matrices
$\boldsymbol{\Theta}=\{\Theta_\nu\}_{\nu=1}^M$, which act as the shift operators over its GNN components. These are rarely known a priori and must be estimated from
data. We are given $S$ realizations
$\{(\mathbf{X}^{(r)},y^{(r)})\}_{r=1}^{S}$, where each
$\mathbf{X}^{(r)}\in\mathbb{R}^{N\times T}$ contains $T$ samples of the process
and $y^{(r)}$ is the associated label, and we denote by
$\tilde{\mathbf{X}}^{(r)}_\nu$ the band-$\nu$ Fourier coefficients of the
$r$-th realization. We propose a \emph{joint} problem formulation which couples statistical estimation of the matrices $\boldsymbol{\Theta}$ by spectral likelihood and the supervised objective, in analogy with the PNN training
of~\cite{cavallo2026pnn}.

\spara{Likelihood objective.}
A common approach in literature for estimating the iCSD matrices is to leverage the \textit{Whittle approximation} \cite{whittle1951,tugnait2022sparse}. For each band, the negative Whittle
pseudo log-likelihood is defined as
\[
    \mathcal{L}_\nu(\Theta_\nu)
    =
    -\log\det(\Theta_\nu)
    + \mathrm{Tr}(K_\nu\,\Theta_\nu),
\]
  \[  K_\nu =
    \frac{1}{S B}\sum_{r=1}^{S}
    \tilde{\mathbf{X}}^{(r)}_\nu\big(\tilde{\mathbf{X}}^{(r)}_\nu\big)^{H},
\]
where $K_\nu$ is the empirical spectral covariance over band $\nu$. Note that averaging across frequencies is justified by the assumption that $\Gamma_f$, and hence $\Theta_f$, is
approximately constant within each band.
Consistently with the conditional-independence structure of the PCGs, we add a
sparse-group lasso penalty. Denoting by
$\theta_{ij}=\big([\Theta_1]_{ij},\dots,[\Theta_M]_{ij}\big)\in\mathbb{C}^{M}$
the entries of a given pair across bands, the penalized loss reads
\[
    \mathcal{L}_{\text{prob}}(\boldsymbol{\Theta})
    =
    \sum_{\nu=1}^{M} \mathcal{L}_\nu(\Theta_\nu)
    + \rho \sum_{i\neq j}
      \Big[
        \alpha\,\|\theta_{ij}\|_1
        + (1-\alpha)\,\|\theta_{ij}\|_2
      \Big],
\]
where $\rho>0$ sets the regularization strength and $\alpha\in[0,1]$
balances band-wise sparsity against edges vanishing across all bands.

\spara{Joint formulation.}
The iCSD operators are learned together with the network parameters, so that
the estimated structure is \textit{aware} of the downstream task. Denoting by $\mathcal{L}_{\text{pred}}(\boldsymbol{\Theta},\mathbf{H})$ the prediction loss of the supervised task, this yields the coupled problem
\begin{equation}
\label{eq:P1}
\tag{P1}
\begin{aligned}
    \min_{\boldsymbol{\Theta},\,\mathbf{H}}
    \quad &
    \beta\,\mathcal{L}_{\text{pred}}(\boldsymbol{\Theta},\mathbf{H})
    +
    (1-\beta)\,\mathcal{L}_{\text{prob}}(\boldsymbol{\Theta})
    \\
    \text{s.t.}
    \quad &
    \Theta_\nu \succ 0,
    \qquad \nu=1,\dots,M,
\end{aligned}
\end{equation}
where $\beta\in[0,1]$ balances the prediction objective against the statistical
fit, and the constraint ensures that each operator remains a valid iCSD matrix. \\
Problem~\eqref{eq:P1} is challenging to address directly. The positive-definiteness
constraint is natural for the likelihood term, which is convex in
$\boldsymbol{\Theta}$, but is impractical to enforce within the prediction loss. We therefore
introduce a relaxed version of the problem, in which we use a set of auxiliary
matrices $\tilde{\boldsymbol{\Theta}}=\{\tilde{\Theta}_\nu\}_{\nu=1}^M$ to decouple the two losses:
\begin{equation}
\label{eq:P2}
\tag{P2}
\begin{aligned}
    \min_{\boldsymbol{\Theta},\,
          \tilde{\boldsymbol{\Theta}},\,
          \mathbf{H}}
    \quad &
    \beta\,\mathcal{L}_{\mathrm{pred}}
    (\boldsymbol{\Theta},\mathbf{H})
    +
    (1-\beta)\,
    \mathcal{L}_{\mathrm{prob}}
    (\tilde{\boldsymbol{\Theta}})
    \\
    &\quad+
    \delta\sum_{\nu=1}^{M}
    \big\|\Theta_\nu-\tilde{\Theta}_\nu\big\|_F^2
    \\
    \text{s.t.}
    \quad &
    \tilde{\Theta}_\nu \succ 0,
    \qquad \nu=1,\dots,M.
\end{aligned}
\end{equation}
Here, $\delta>0$ controls the consistency between the two sets of matrices, and
as $\delta\to\infty$ the constraint $\Theta_\nu=\tilde{\Theta}_\nu$ is enforced
exactly, recovering~\eqref{eq:P1}.

%% file: sections/problem_solution.tex
\section{Problem Solution}
\label{sec:algorithm}
We address parameter estimation of the iCSNN  by block coordinate descent, exploiting the decoupled problem~\eqref{eq:P2}. Since both the operators and the GNN parameters are complex-valued, their
gradient-based updates are computed through Wirtinger
calculus~\cite{Schreier2010}.
For $I_{\max}$ iterations, the three sets of variables are optimized in turn, according to the following
update rules.

\spara{Likelihood update ($\tilde{\boldsymbol{\Theta}}$).}
With $\boldsymbol{\Theta}$ fixed, the auxiliary matrices solve the penalized
spectral likelihood problem augmented with the coupling term,
\[
    \tilde{\boldsymbol{\Theta}}^{(t+1)}
    =
    \arg\min_{\tilde{\Theta} \succ 0}
    \;(1-\beta)\,\mathcal{L}_{\mathrm{prob}}(\tilde{\boldsymbol{\Theta}})
    + \delta\,\sum_{\nu=1}^{M}
      \big\|\Theta_\nu^{(t)} - \tilde{\Theta}_\nu\big\|_F^2  .
\]
This problem is convex and can be solved following a similar approach
to~\cite{tugnait2022sparse}. In practice, we optimize it by ADMM for $I_{\tilde{\boldsymbol{\Theta}}}$ iterations, alternating between a likelihood
step and a sparse-group lasso proximal step, available in closed form. The consistency term only adds a further quadratic contribution to the
likelihood step, which is still solved in closed form, along the same lines
as~\cite{tugnait2022sparse}.  

\spara{Operator update ($\boldsymbol{\Theta}$).}
With $\mathbf{H}$ and $\tilde{\boldsymbol{\Theta}}$ fixed, the operators
entering the network are refined in the supervised problem, subject to the
coupling term,
\[
    \boldsymbol\Theta^{(t+1)}
    =
    \arg\min_{\boldsymbol{\Theta}}
    \;\beta\,\mathcal{L}_{\mathrm{pred}}(\boldsymbol{\Theta},\mathbf{H}^{(t)})
    + \delta\,\sum_{\nu=1}^{M}
      \big\|\Theta_\nu - \tilde{\Theta}^{(t+1)}_\nu\big\|_F^2 .
\]
This subproblem is unconstrained, and, for $\delta$ large enough, the quadratic term makes it locally
strongly convex, which we exploit by L-BFGS optimization for $I_{\boldsymbol{\Theta}}$ iterations \cite{CVX}.

\spara{Network update ($\mathbf{H}$).}
Finally, with the operators fixed, the network parameters minimize the
prediction loss alone,
\[
    \mathbf{H}^{(t+1)}
    =
    \arg\min_{\mathbf{H}}
    \;\mathcal{L}_{\mathrm{pred}}(\boldsymbol{\Theta}^{(t+1)},\mathbf{H}),
\]
which is solved by stochastic gradient descent for $I_{\mathbf{H}}$ iterations.

%% file: sections/numerical_results.tex
\begin{figure}
    \centering
    \includegraphics[width = .8\columnwidth]{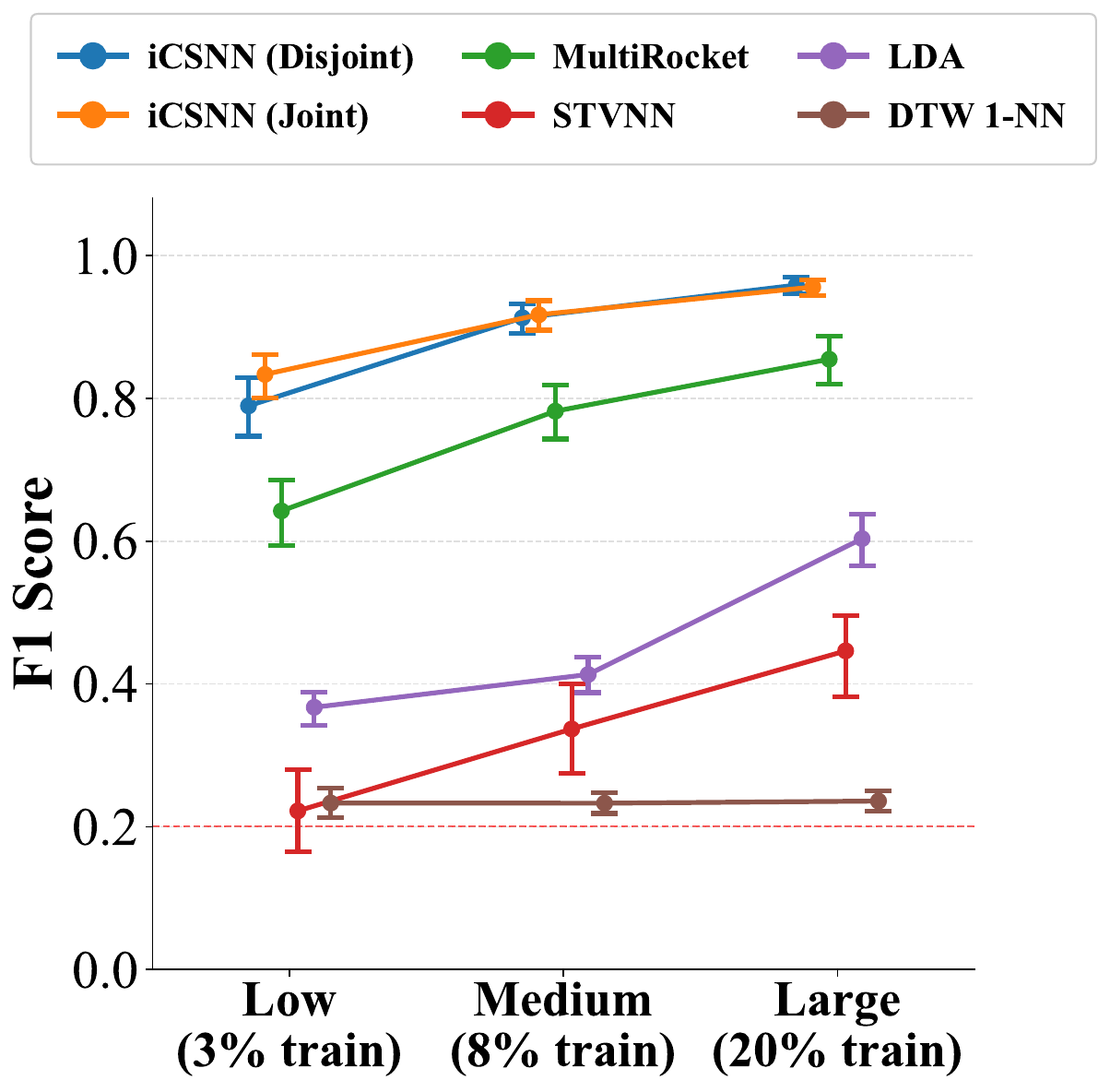}
\caption{Macro-averaged F1 scores of the jointly trained iCSNN, its Disjoint ablation and baseline methods, averaged over 20 synthetic datasets. 
The red dashed line represents the chance-level score of $0.20$ for five balanced classes.} 
    \label{F1_score}
\end{figure}

\section{Numerical Results}
We evaluate iCSNNs on a synthetic multivariate time-series classification
task. 

\spara{Data generation.} A dataset consists of finite-length realizations from $C=5$ distinct
stationary processes, corresponding to five classes. Each realization is a
multivariate time series with $N=10$ channels and length $T=100$. We generate $1000$ independent realizations for each process, yielding a total of
$5000$ samples. For $i=1,\ldots,C$, class $i$ corresponds to a latent state that determines the dynamics of an informative component, which is combined with a state-independent component:
\[
    X^{(i)}(t)
    =
    B_s z_s^{(i)}(t) + B_c z_c^{(i)}(t)
    \in \mathbb{R}^N,
\]
where $[B_s \mid B_c]\in\mathbb{R}^{N\times N}$ is a random orthogonal mixing
matrix, with $B_s^\top B_c=0$. The latent component
$z_s^{(i)}(t)$ is a class-specific zero-mean Gaussian VAR($3$) process on
$d_s=3$ channels and contains the discriminative information, whereas
$z_c^{(i)}(t)$ is a zero-mean Gaussian VAR($3$) process on the remaining
$7$ channels, with parameters shared across classes, and acts as shared noise.
Hence, the classes differ only through the second-order
structure of a low-dimensional latent component, which is mixed with a
higher-dimensional class-independent process. 
To assess performance under different data availability regimes, we consider train/validation/test splits in ${(3\%,3\%,94\%),(8\%,8\%,84\%),(20\%,20\%,60\%)}$, corresponding to low-, medium-, and large-data settings, respectively.

\spara{Baselines.} We compare iCSNN against five baselines from different families of methods. The first, denoted \emph{Disjoint}, is an ablation of our method in
which the iCSD matrices are estimated beforehand from the spectral likelihood
alone. 
This isolates the contribution of the task-driven refinement of the operators. To compare
against existing covariance-based architectures, we consider STVNN \cite{Cavallo_2024}, which uses the marginal covariance
$C=\mathbb{E}[X(t)X(t)^\top]$ as graph shift operator. Since STVNNs were originally designed for online forecasting, we adapt them to classification by using a covariance shift operator estimated once from the training set, and by replacing the forecasting head with a graph-level classification readout followed by an MLP. The comparison is not entirely on equal terms, as STVNNs were developed for online optimization and forecasting rather than classification. Nevertheless, we include this baseline as the closest covariance-based approach to our method. The graph filtering architecture is otherwise unchanged. We further include Linear Discriminant Analysis (LDA) on per-node log-periodogram
features,
\[
    \phi(\mathbf{X}) =
    \Big[
    \log\!\big(1+|\tilde{X}^{(n)}_f|^2\big)
    \Big]_{n=1,\dots,N;\ f=1,\dots,F}
    \in \mathbb{R}^{NF},
\]
which captures marginal spectral information without cross-node dependencies. 
Finally, we include MultiRocket~\cite{tan2022multirocket}, a state-of-the-art time-series
classifier, which extracts features from fixed random convolutional kernels, and a $1$-nearest-neighbour classifier with
Dynamic Time Warping distance $d_{\mathrm{DTW}}$ as a distance-based reference \cite{ding2008querying}.

\spara{Training.}
We train an iCSNN composed of $M=8$ branches, each a GNN with two iCSD
filtering layers of order $K=2$, followed by an MLP head with hidden sizes
$\{16,8\}$. The architecture is trained by solving~\eqref{eq:P2} with
regularization and block regularization strengths $\alpha=0.1$ and $\rho=10$, respectively,
$\beta=0.5$ balancing the prediction and likelihood objectives, and the consistency term $\delta=0.1$. The hyperparameters and learning rates of all methods are selected on a tuning dataset with the Optuna optimization module \cite{akiba2019optuna}. We instantiate the algorithm of Section~\ref{sec:algorithm} with $I_{\max}=5$
outer iterations. The likelihood and network updates are run for $I_{\tilde{\boldsymbol{\Theta}}}=I_{\mathbf{H}}=500$ iterations each, whereas the
operator update requires only $I_{\boldsymbol{\Theta}}=30$, since L-BFGS is expected to converge in few iterations in a locally strongly convex setting.

 \spara{Results.} We evaluate iCSNN and the considered baselines over $20$ datasets, randomly generated according to the described setting. We evaluate performance using the
macro-averaged F1 score,
\[
    F1_{\mathrm{macro}}
    =
    \frac{1}{C}
    \sum_{i=1}^{C}
    \frac{2P_iR_i}{P_i+R_i},
\]
where $P_i$ and $R_i$ denote the precision and recall of class $i$, respectively. 
Fig.~\ref{F1_score} reports the mean macro-F1 score across the $20$ simulations, together with $95\%$ confidence intervals. Both iCSNN variants consistently outperform the remaining baselines across
all data regimes. In the low-data setting, the jointly trained iCSNN achieves the highest mean F1 score, improving over its \emph{Disjoint} ablation. As the amount of training data increases, the performance of both variants improves
and the gap between them progressively vanishes, with nearly identical results in the large-data regime. MultiRocket also benefits substantially from additional training data, but remains below the iCSNN variants throughout. LDA and STVNN exhibit a marked improvement from the low- to the large-data
regime, although their performance remains considerably lower, whereas DTW $1$-NN stays close to the chance-level score of $0.20$ across all settings.

The results reflect the design of the synthetic task,  where
class information is encoded in the second-order temporal and cross-variable
dependence structure. LDA uses
only marginal spectral features, MultiRocket does not explicitly model conditional cross-node structure, and DTW $1$-NN ignores the spectral dependence representation. Similarly, STVNN relies on a static covariance operator, which marginalizes temporal information and does not explicitly encode the
frequency-specific structures separating the classes. Its performance should also be viewed in light of its adaptation from online forecasting to offline classification. In contrast, iCSNN and \emph{Disjoint} exploit bandwise iCSD operators, accounting for temporal correlations through the frequency-domain representation and for cross-channel conditional dependencies through the spectral precision
operators. The gain of iCSNN over \emph{Disjoint} further suggests that task-driven refinement of the iCSD is beneficial in the low-data regime, where
likelihood-based estimates can be highly variable and may induce
less stable representation spaces.

%% file: sections/conclusions.tex
\section{Conclusions}

In this work, we introduced Inverse Cross-Spectral Neural Networks (iCSNNs), a class of GNNs that extends covariance-based graph neural networks for multivariate time-series analysis. Under stationarity, iCSNNs exploit the spectral representation to factorize second-order dependencies across frequencies, with bandwise iCSD operators encoding frequency-specific conditional dependencies among variables. This yields a compact architecture that preserves the process dependence structure with a parsimonious parametrization. We also proposed a joint learning procedure that combines spectral likelihood-based estimation with task-driven refinement of the iCSD operators and training of the network parameters. Numerical experiments on a synthetic classification task, in which class information is encoded in the second-order cross-spectral structure, show that iCSNNs outperform baselines from different families of methods across considered data regimes. The jointly trained iCSNN further improves over a two-stage variant in which the iCSD operators are estimated from the spectral likelihood alone in the low-data regime, indicating that task-aware learning of the operators is most beneficial when likelihood-based spectral estimates are highly variable.

A promising research direction is to investigate whether iCSNNs trained on time series of length $T$ can generalize to sequences of different lengths (length-transferability). This is motivated by the
fact that their spectral filtering operators describe second-order properties of the underlying stationary process rather than a specific observation horizon. This could support predictions that remain consistent across recording lengths in tasks where the target reflects properties of the generating process, with longer observations primarily improving prediction reliability. 
Finally, we will assess the proposed architecture on
real-world multivariate time-series datasets, where interpretable modeling of
cross-variable dependencies is an important requirement.

%% file: bibliography.bib
@article{bullmore2009complex,
  author  = {Bullmore, E. and Sporns, O.},
  title   = {Complex brain networks: Graph theoretical analysis of structural and functional systems},
  journal = {Nature Reviews Neuroscience},
  volume  = {10},
  number  = {3},
  pages   = {186--198},
  year    = {2009},
  doi     = {10.1038/nrn2575}
}

@inproceedings{akiba2019optuna,
  author    = {Akiba, T. and Sano, S. and Yanase, T. and
               Ohta, T. and Koyama, M.},
  title     = {Optuna: A Next-generation Hyperparameter Optimization Framework},
  booktitle = {Proceedings of the 25th ACM SIGKDD International Conference on
               Knowledge Discovery \& Data Mining},
  pages     = {2623--2631},
  year      = {2019},
  doi       = {10.1145/3292500.3330701},
  url       = {https://doi.org/10.1145/3292500.3330701}
}

@book{CVX, 
  title={Convex optimization}, 
  author={S. Boyd and L. Vandenberghe}, 
  year={2004}, 
  publisher={Cambridge {U}niversity {P}ress} 
}

@ARTICLE{Bach2004,
  author={Bach, F.R. and Jordan, M.I.},
  journal={IEEE Transactions on Signal Processing}, 
  title={Learning graphical models for stationary time series}, 
  year={2004},
  volume={52},
  number={8},
  pages={2189-2199},
  doi={10.1109/TSP.2004.831032}}

@article{dahlhaus2000graphical,
  author  = {Dahlhaus, R.},
  title   = {Graphical interaction models for multivariate time series},
  journal = {Metrika},
  volume  = {51},
  pages   = {157--172},
  year    = {2000},
  doi     = {10.1007/s001840000055}
}

@book{whittle1951,
  title={Hypothesis Testing in Time Series Analysis},
  author={Whittle, P.},
  year={1951},
  publisher={Almqvist \& Wiksell},
  address={Uppsala}
}

@ARTICLE{dacunto2024,
  author={D’Acunto, G. and Di Lorenzo, P. and Bonchi, F. and Sardellitti, S. and Barbarossa, S.},
  journal={IEEE Transactions on Signal Processing}, 
  title={Learning Multi-Frequency Partial Correlation Graphs}, 
  year={2024},
  volume={72},
  number={},
  pages={2953-2969},
  doi={10.1109/TSP.2024.3401072}}

@article{Schreier2010,
    author = { Schreier, P. J. and Scharf, L. L.},
    title = {Statistical signal processing of complex-valued data : the theory of improper and noncircular signals},
    journal = {Cambridge University Press},
    year = 2010
}

@inbook{Cavallo_2024,
   title={Spatiotemporal Covariance Neural Networks},
   ISBN={9783031703447},
   ISSN={1611-3349},
   url={http://dx.doi.org/10.1007/978-3-031-70344-7_2},
   DOI={10.1007/978-3-031-70344-7_2},
   booktitle={Machine Learning and Knowledge Discovery in Databases. Research Track},
   publisher={Springer Nature Switzerland},
   author={Cavallo, A. and Sabbaqi, M. and Isufi, E.},
   year={2024},
   pages={18–34} }

@book{brockwell1991time,
  author    = {Brockwell, P. J. and Davis, R. A.},
  title     = {Time Series: Theory and Methods},
  edition   = {2},
  publisher = {Springer},
  year      = {1991},
  doi       = {10.1007/978-1-4419-0320-4}
}

@article{tugnait2022sparse,
  author  = {Tugnait, J. K.},
  title   = {On sparse high-dimensional graphical model learning for dependent time series},
  journal = {Signal Processing},
  volume  = {197},
  pages   = {108539},
  year    = {2022},
  doi     = {10.1016/j.sigpro.2022.108539}
}

@inproceedings{cavallo2026pnn,
  author    = {Cavallo, A. and Rey, S. and Marques, A. G. and Isufi, E.},
  title     = {Precision Neural Networks: Joint graph and relational learning},
  booktitle = {Proc. IEEE Int. Conf. Acoust., Speech, Signal Process. (ICASSP)},
  address   = {Barcelona, Spain},
  year      = {2026}
}

@ARTICLE{transferability_VNN,
  author={Sihag, S. and Mateos, G. and McMillan, C. and Ribeiro, A.},
  journal={IEEE Journal of Selected Topics in Signal Processing}, 
  title={Transferability of coVariance Neural Networks}, 
  year={2024},
  volume={18},
  number={2},
  pages={199-215},
  doi={10.1109/JSTSP.2024.3378887}}

@ARTICLE{GNN_on_timeseries,
author={Jin, M. and Koh, H. Y. and Wen, Q. and Zambon, D. and Alippi, C. and Webb, G. I. and King, I. and Pan, S.},
journal={ IEEE Transactions on Pattern Analysis \& Machine Intelligence },
title={{ A Survey on Graph Neural Networks for Time Series: Forecasting, Classification, Imputation, and Anomaly Detection }},
year={2024},
volume={46},
number={12},
ISSN={1939-3539},
pages={10466-10485},
doi={10.1109/TPAMI.2024.3443141},
url = {https://doi.ieeecomputersociety.org/10.1109/TPAMI.2024.3443141},
publisher={IEEE Computer Society},
address={Los Alamitos, CA, USA},
month=dec}

@article{tan2022multirocket,
  author  = {Tan, C. W. and Dempster, A. and Bergmeir, C. and Webb, G. I.},
  title   = {{MultiRocket}: Multiple pooling operators and transformations for fast and effective time series classification},
  journal = {Data Mining and Knowledge Discovery},
  volume  = {36},
  number  = {5},
  pages   = {1623--1646},
  year    = {2022},
  doi     = {10.1007/s10618-022-00844-1}
}

@article{ding2008querying,
  author    = {Ding, H. and Trajcevski, G. and Scheuermann, P. and Wang, X. and Keogh, E.},
  title     = {Querying and mining of time series data: Experimental comparison of representations and distance measures},
  journal   = {Proceedings of the VLDB Endowment},
  volume    = {1},
  number    = {2},
  pages     = {1542--1552},
  year      = {2008},
  doi       = {10.14778/1454159.1454226}
}

@book{tsay2013multivariate,
  title     = {Multivariate Time Series Analysis: With R and Financial Applications},
  author    = {Tsay, R. S.},
  year      = {2013},
  publisher = {John Wiley \& Sons},
  address   = {Hoboken, NJ},
  isbn      = {978-1-118-61790-8}
}

@article{Runge2019,
  author = {Runge, J. and Bathiany, S. and Bollt, E. and Camps-Valls, G. and Coumou, D. and Deyle, E. and Glymour, C. and Kretschmer, M. and Mahecha, M. D. and Muñoz-Marí, J. and Others},
  title = {Inferring causation from time series in Earth system sciences},
  journal = {Nature Communications},
  volume = {10},
  number = {2553},
  year = {2019},
  doi = {10.1038/s41467-019-10105-3},
  url = {https://www.nature.com/articles/s41467-019-10105-3}
}

@inproceedings{VNNs,
 author = {Sihag, S. and Mateos, G. and McMillan, C. and Ribeiro, A.},
 booktitle = {Advances in Neural Information Processing Systems},
 doi = {10.52202/068431-1237},
 editor = {S. Koyejo and S. Mohamed and A. Agarwal and D. Belgrave and K. Cho and A. Oh},
 pages = {17003--17016},
 publisher = {Curran Associates, Inc.},
 title = {coVariance Neural Networks},
 url = {https://proceedings.neurips.cc/paper_files/paper/2022/file/6cb00ce1a21a090a3dae04cebebd8341-Paper-Conference.pdf},
 volume = {35},
 year = {2022}
}
